\documentclass[preprint,3p,times,authoryear]{elsarticle}

\usepackage{amsmath,amssymb}
\usepackage{microtype}
\usepackage{booktabs}
\usepackage{tabularx}
\usepackage{array}
\usepackage{graphicx}
\usepackage{adjustbox}
\usepackage{subcaption}
\usepackage{url}
\usepackage[hidelinks]{hyperref}
\usepackage{orcidlink}
\usepackage{tikz}
\usepackage{helvet}
\usetikzlibrary{arrows.meta,backgrounds,fit,decorations.pathmorphing}

\journal{Computers and Education: Artificial Intelligence}

\begin{document}

\begin{frontmatter}

\title{Automated Grading of Handwritten Assessments with Multimodal Large Language Models: Methodology and Empirical Evaluation}

\author[a]{Pierre Schaus\,\orcidlink{0000-0002-3153-8941}\corref{cor1}}
\ead{pierre.schaus@uclouvain.be}

\author[a]{Beno\^{i}t Ronval\,\orcidlink{0009-0008-1163-8831}}
\ead{benoit.ronval@uclouvain.be}

\cortext[cor1]{Corresponding author.}

\address[a]{ICTEAM Institute, Universit\'{e} catholique de Louvain (UCLouvain), Place Sainte Barbe 2, bte L5.02.01, 1348 Louvain-la-Neuve, Belgium}

\begin{abstract}
While the proliferation of generative artificial intelligence has prompted many institutions to return to supervised, pen-and-paper examinations to ensure academic integrity, grading large cohorts of handwritten scripts remains a severe bottleneck. Manual marking is time-consuming, vulnerable to inter-rater and intra-rater noise and cognitive biases, and rarely leaves instructors time to write individualized formative feedback beyond bare rubric labels. In this paper, we present and empirically evaluate a stateless multimodal Large Language Model (LLM) grading methodology for open-ended handwritten assessments. For each question, the pipeline evaluates a cropped image of the student's handwritten response against explicit, strictly additive rubric criteria in a single model invocation—outputting Boolean criterion selections, an auditable justification for the examiner, and personalized formative feedback for the student—while computing the numerical score deterministically outside the LLM. We evaluate this approach on a high-stakes civil engineering entrance examination in mathematics comprising $1{,}752$ handwritten copies across four subjects (Geometry, Trigonometry, Analysis, and Algebra) previously graded by human instructors. Across five research questions, we show that: (1) frontier multimodal grading aligns closely with human scores (Pearson correlation $r$ between $0.92$ and $0.97$, and mean absolute error between $4.7\%$ and $7.1\%$ of the exam scale); (2) a frontier model substantially reduces leniency bias and dispersion compared to a compact open-weights alternative on multi-step symbolic derivations; (3) the generated feedback is genuinely copy-specific, enabling a blind judge to match a feedback paragraph back to its target script among five candidates in $90\%$ of cases; (4) all $94$ rubric criteria are meaningfully utilized across the cohort; and (5) grading is highly repeatable across independent runs, achieving a single-run intraclass correlation of $0.99$ and a mean pairwise score difference of $1.4\%$ of the scale.
\end{abstract}


\end{frontmatter}

\section{Introduction}

Grading is a fundamental pillar of education, serving both as a summative measure of student mastery and as a vehicle for formative feedback that guides learning \citep{black1998assessment, sadler1989formative, shute2008focus}.
In recent years, the ubiquitous availability of conversational Generative Artificial Intelligence (GenAI) has made it increasingly difficult for universities to trust unproctored take-home projects, programming assignments, or homework as authentic reflections of student understanding \citep{baidoo2023education, kasneci2023chatgpt, cotton2024chatting}.
Consequently, institutions have shifted weight back toward supervised, in-person assessments.
Because oral interviews cannot scale to large cohorts, educators typically face a trade-off between Multiple-Choice Questions (MCQs) and open-ended pen-and-paper examinations.
While MCQs automate grading, they cannot directly evaluate multi-step mathematical derivations, proof construction, or graphical sketching, are often disfavored by students when minor slips at the end of a long calculation yield zero credit, and introduce psychometric biases such as \textit{test-wiseness} \citep{millman1965analysis} and differential risk-taking under negative or threshold scoring \citep{baldiga2014gender}.
Open-ended handwritten examinations avoid these pitfalls by directly assessing students' reasoning processes, but grading them manually at scale creates a severe operational bottleneck that is structurally vulnerable to the two fundamental flaws of human judgment formalized by \citet{kahneman2021noise}: systematic \textit{bias} and unwanted \textit{noise}.

The pervasiveness of noise in examination marking has been documented since the landmark studies of \citet{starch1913reliability} and \citet{pieron1963examens}, who showed that independent examiners grading the exact same mathematics or Baccalaur\'{e}at scripts exhibit massive score dispersion.
In the taxonomy of \citet{kahneman2021noise}, multi-grader marking suffers from both \textit{level noise} (baseline differences in grader severity or leniency) and \textit{pattern noise} (idiosyncratic reactions to specific solution strategies or errors).
Even when a single instructor grades an entire cohort—eliminating inter-rater level noise—human evaluation remains heavily distorted by intra-rater \textit{occasion noise}, whereby the same evaluator assigns different scores to equivalent work depending on when and in what sequence it is inspected.
Prolonged grading sessions induce cognitive overload and rater fatigue \citep{sweller1988cognitive}, causing severity to drift over time \citep{mahshanian2020effect}.
Moreover, sequential grading stacks expose evaluators to strong \textit{contrast effects}: an average script receives significantly higher marks when graded immediately after weak scripts and lower marks after strong ones \citep{daly1982contrast, spear1997influence}.
Compounding this stochastic and sequential variability are systematic cognitive biases \citep{malouff2016bias}, such as halo effects and handwriting bias, where messy or less legible handwriting significantly depresses content scores even when graders are instructed to evaluate substance alone \citep{briggs1980study, klein2005effect}.

To improve fairness and consistency, modern assessment workflows rely on analytic scoring rubrics \citep{andrade2005teaching, panadero2013use} and digital platforms such as Gradescope\footnote{\url{https://www.gradescope.com/}} \citep{singh2017gradescope}.
From the students' perspective, rubric-based grading is widely appreciated because explicit criteria are perceived as transparent and objective \citep{andrade2005teaching, panadero2013use}.
However, reviews of the rubric literature and empirical studies of assessors show that written rubric descriptors remain ambiguous without intensive rater calibration \citep{jonsson2007use, brookhart2015quality, bloxham2016let}, and human graders frequently form an immediate holistic impression before reverse-engineering criterion selections to match it \citep{sadler2009indeterminacy}.
Furthermore, because platforms like Gradescope do not automate the evaluation itself, instructors must still manually inspect every copy and click criteria one by one.
Standard rubric labels remain generic and succinct: they tell a student \textit{what} criterion was missed, but rarely explain \textit{why} their specific handwritten derivation failed or \textit{how} to correct their reasoning \citep{shute2008focus}.
\\

Prior efforts to automate the grading of open-ended assessments have evolved through three main paradigms \citep{burrows2015eras}.
Early Automated Short-Answer Grading (ASAG) systems relied on lexical, semantic-similarity, or rule-based text matching \citep{leacock2003c, burrows2015eras}, restricting their applicability to short typed text rather than handwritten multi-step derivations.
Subsequent pipelines for handwritten STEM assessments adopted a two-stage architecture combining optical character recognition (OCR) or mathematical expression recognition with a downstream symbolic or text evaluator \citep{kortemeyer2024grading}; however, transcription errors in dense or crossed-out handwriting propagate directly to the grading stage, and text-only OCR pipelines cannot evaluate geometric drawings, sign tables, or spatial layouts.
More recently, Large Language Models (LLMs) have demonstrated strong capabilities in complex reasoning tasks \citep{zhao2026survey, ferrag2026llm}, and multimodal Vision-Language Models extend these capabilities to directly process handwritten images, mathematical notation, and diagrams alongside text \citep{xie2025large, yan2025survey}.
Yet asking a language model to directly output a holistic numerical score for an exam script remains vulnerable to arithmetic summing errors, uncalibrated scoring drift, and opaque reasoning \citep{kortemeyer2024grading}.
Combining multimodal Vision-Language Models with explicit criterion-level decomposition offers the possibility of uniting the scalability of MCQs with the transparency of analytic rubrics, while going a step further: generating individualized formative feedback tailored to each student's script.
\\

To investigate this potential, we formalize a stateless, criterion-based multimodal LLM grading methodology (Section~\ref{sec:approach}) and conduct a large-scale empirical evaluation of its accuracy, feedback specificity, and reliability on a high-stakes, real-world benchmark: the June 2026 session of the mandatory mathematics entrance examination for civil engineering studies in Wallonia (Belgium), organized at UCLouvain.
Our dataset comprises $1{,}752$ handwritten copies across four distinct examinations—Geometry, Trigonometry, Analysis, and Algebra (435 to 442 copies per subject)—featuring symbolic calculations, multi-step proofs, tables, and geometric drawings, all previously graded by teams of human instructors on Gradescope.
To compare AI grading directly against operational human Gradescope evaluations, we retained the exact question point weightings and reformulated the instructors' Gradescope rubrics into explicit additive criteria calibrated on at most 4 to 5 sample copies per question.
Through this experimental study, we investigate five core research questions, ordered as they are addressed in Section~\ref{sec:results}:
\begin{itemize}
    \item \textbf{RQ1 (Alignment with Human Grading):} Are automated evaluations produced by multimodal LLMs consistent with those of expert human graders across different mathematical subjects and question types?
    \item \textbf{RQ2 (Open-Weights vs.\ Frontier Models):} Are open-weights multimodal LLMs competitive with proprietary frontier models in grading accuracy?
    \item \textbf{RQ3 (Generated Feedback):} Are the generated feedbacks specific to each of the evaluated questions and students?
    \item \textbf{RQ4 (Criterion Utilization):} Are all defined rubric criteria effectively and meaningfully utilized by the model across the student cohort?
    \item \textbf{RQ5 (Run-to-Run Stability):} To what extent does stochasticity in LLM inference cause score variations from one grading run to another on the exact same copies?
\end{itemize}

Our results show that multimodal AI grading strongly aligns with human grading—reaching a Pearson correlation between $0.92$ and $0.97$ across the four exams with \texttt{gemini-3.8-flash}—while outperforming compact open-weights alternatives on multi-step derivations, generating feedback specific enough that an LLM judge identifies the target copy among five candidates in $90\%$ of cases, utilizing the full spectrum of rubric criteria, and exhibiting high run-to-run stability (a mean difference of $0.58$ points on a $40$-point scale across independent runs).

The remainder of this paper is structured as follows: Section~\ref{sec:approach} details the global multimodal grading methodology and pipeline.
Section~\ref{sec:setup} describes the benchmark datasets and evaluated multimodal LLMs, and Section~\ref{sec:results} presents and analyzes the empirical results for RQ1--RQ5.
Finally, Section~\ref{sec:limitations} outlines limitations and future work, and Section~\ref{sec:conclusion} concludes the paper.

\section{Automated Grading Methodology}
\label{sec:approach}

\begin{figure}[t]
    \centering
    \resizebox{\textwidth}{!}{%
        \begin{tikzpicture}[font=\scriptsize, >=Stealth]
  \definecolor{inBg}{HTML}{F8FAFC}      \definecolor{inHdr}{HTML}{E2E8F0}
  \definecolor{inBdr}{HTML}{1E3A8A}     \definecolor{inTxt}{HTML}{1E3A8A}
  \definecolor{llmBg}{HTML}{F1F5F9}     \definecolor{llmHdr}{HTML}{CBD5E1}
  \definecolor{llmBdr}{HTML}{1E3A8A}    \definecolor{llmTxt}{HTML}{1E3A8A}
  \definecolor{jsonBg}{HTML}{F0FDFA}    \definecolor{jsonHdr}{HTML}{CCFBF1}
  \definecolor{jsonBdr}{HTML}{0D9488}   \definecolor{jsonTxt}{HTML}{115E59}
  \definecolor{sumBg}{HTML}{FFFBEB}     \definecolor{sumHdr}{HTML}{FDE68A}
  \definecolor{sumBdr}{HTML}{D97706}    \definecolor{sumTxt}{HTML}{92400E}
  \definecolor{passCol}{HTML}{115E59}   \definecolor{failCol}{HTML}{B91C1C}
  \definecolor{mutedCol}{HTML}{374151}  \definecolor{arrowCol}{HTML}{1F2937}

  \tikzset{
    boxpanel/.style={
      rectangle, rounded corners=3pt, draw=#1Bdr, line width=0.65pt,
      fill=white, inner xsep=6pt, inner ysep=5pt, align=left
    },
    stagebox/.style={
      rectangle, rounded corners=4pt, draw=#1Bdr, line width=0.85pt,
      fill=#1Bg, inner sep=6pt
    },
    stagetitle/.style={
      font=\footnotesize\bfseries, text=#1Txt, align=center
    },
    flow/.style={
      draw=arrowCol, line width=0.85pt, -{Stealth[length=5pt, width=4.5pt]}
    },
    dashflow/.style={
      draw=sumBdr, line width=0.8pt, densely dashed, -{Stealth[length=4.5pt, width=4pt]}
    }
  }

  \node[boxpanel=in, text width=132pt] (img) at (0, 0) {%
    {\color{inTxt}\bfseries 1. Cropped Question Image}\\[2pt]
    {\color{mutedCol}Isolated bounding box from scan:}\\[3pt]
    {\setlength{\fboxsep}{4pt}\setlength{\fboxrule}{0.45pt}%
     \fcolorbox{inBdr!70}{inBg}{%
       \parbox{\dimexpr\linewidth-8pt-1pt\relax}{%
         {\tiny\bfseries\color{inTxt} Q1: Solve $\sin^2(4x) + \cos^4(2x) = 1$}\\[1pt]
         {\color{inBdr!50}\hrule height 0.35pt}\vspace{2pt}
         {\tiny\itshape\color{llmTxt} $\sin(4x)=2\sin(2x)\cos(2x) \implies \dots$}\\[1pt]
         {\tiny\itshape\color{llmTxt} $\sin^2(2x)[3\cos^2(2x)-1]=0 \implies \dots$}%
       }%
     }}%
  };

  \node[boxpanel=in, text width=132pt, anchor=north] (crit) at ([yshift=-5pt]img.south) {%
    {\color{inTxt}\bfseries 2. Additive Rubric Criteria $\mathcal{C}_q$}\\[2pt]
    \textbf{C1} {\color{sumTxt}\textbf{(+2.0 pt)}}: ``Double-angle step''\\
    \textbf{C2} {\color{sumTxt}\textbf{(+2.0 pt)}}: ``Factorization step''\\
    \textbf{C3} {\color{sumTxt}\textbf{(+1.5 pt)}}: ``Solve branch 1''\\
    \textbf{C4} {\color{sumTxt}\textbf{(+2.0 pt)}}: ``Solve branch 2 ($\pm$)''\\
    \textbf{C5} {\color{sumTxt}\textbf{(+0.5 pt)}}: ``Solution set \& $k\in\mathbb{Z}$''%
  };

  \node[boxpanel=in, text width=132pt, anchor=north] (instr) at ([yshift=-5pt]crit.south) {%
    {\color{inTxt}\bfseries 3. Grading \& Feedback Instructions}\\[2pt]
    {\color{mutedCol}\itshape ``Evaluate each criterion independently; accept equivalent methods; explain any error constructively in LaTeX\dots''}%
  };

  \node[stagetitle=in, anchor=south] (s1title) at ([yshift=3pt]img.north) {Stage 1: Question-Level Inputs};
  \begin{scope}[on background layer]
    \node[stagebox=in, fit=(s1title)(img)(crit)(instr)] (stage1) {};
  \end{scope}

  \node[stagebox=llm, text width=72pt, minimum height=100pt, align=center, anchor=west] (llm)
    at ([xshift=22pt, yshift=8pt]stage1.east) {};
  \node[stagetitle=llm, anchor=north] at ([yshift=-6pt]llm.north) {Stage 2:\\Multimodal\\LLM};

  \begin{scope}[shift={([yshift=-3pt]llm.center)}, llmBdr]
    \foreach \y in {-8, 0, 8} {
      \foreach \yy in {-10, -3, 4, 11} {
        \draw[line width=0.35pt, llmBdr!60] (-12pt, \y pt) -- (0pt, \yy pt);
        \draw[line width=0.35pt, llmBdr!60] (0pt, \yy pt) -- (12pt, \y pt);
      }
    }
    \foreach \y in {-8, 0, 8} {
      \filldraw[fill=white, draw=llmBdr, line width=0.55pt] (-12pt, \y pt) circle (1.8pt);
      \filldraw[fill=white, draw=llmBdr, line width=0.55pt] (12pt, \y pt) circle (1.8pt);
    }
    \foreach \yy in {-10, -3, 4, 11} {
      \filldraw[fill=llmHdr, draw=llmBdr, line width=0.55pt] (0pt, \yy pt) circle (1.8pt);
    }
  \end{scope}
  \node[font=\tiny, text=llmTxt, align=center, anchor=south]
    at ([yshift=4pt]llm.south) {Single Stateless Call\\(Schema-Constrained)};

  \node[boxpanel=json, text width=156pt, anchor=north west] (sel)
    at ([xshift=28pt, yshift=-17pt]llm.east |- stage1.north) {%
    {\color{jsonTxt}\bfseries A. Boolean Criteria Selection (\texttt{true}/\texttt{false})}\\[2pt]
    \texttt{\textbf{C1}: \textcolor{passCol}{true}} \quad
    \texttt{\textbf{C2}: \textcolor{passCol}{true}} \quad
    \texttt{\textbf{C3}: \textcolor{passCol}{true}}\\[1pt]
    \texttt{\textbf{C4}: \textcolor{failCol}{false}} \quad
    \texttt{\textbf{C5}: \textcolor{passCol}{true}}%
  };

  \node[boxpanel=json, text width=156pt, anchor=north] (just) at ([yshift=-5pt]sel.south) {%
    {\color{jsonTxt}\bfseries B. Criterion Justification (for Examiner)}\\[2pt]
    {\color{mutedCol}``\textbf{C1--C3} met. In \textbf{C4}, student wrote $\cos(2x)=\frac{\sqrt{3}}{3}$, omitting the negative root $-\frac{\sqrt{3}}{3}$. \textbf{C5} met.''}%
  };

  \node[boxpanel=json, text width=156pt, anchor=north] (fb) at ([yshift=-5pt]just.south) {%
    {\color{jsonTxt}\bfseries C. Personalized Feedback (for Student)}\\[2pt]
    {\color{mutedCol}\itshape ``Clear factorization and Case~1 solution. In Case~2, remember to include both signs $\pm\frac{\sqrt{3}}{3}$ when taking the square root of $\cos^2(2x)=\frac{1}{3}$ \dots''}%
  };

  \node[stagetitle=json, anchor=south] (s3title) at ([yshift=3pt]sel.north) {Stage 3: Structured JSON Output};
  \begin{scope}[on background layer]
    \node[stagebox=json, fit=(s3title)(sel)(just)(fb)] (stage3) {};
  \end{scope}

  \node[stagebox=sum, text width=160pt, align=center, anchor=north] (stage4)
    at ([yshift=-14pt]stage3.south) {%
    {\color{sumTxt}\footnotesize\bfseries Stage 4: Deterministic Score Aggregation}\\[2pt]
    $S(q) = \sum_{i=1}^{k} w_i \cdot \mathbb{I}[\text{criterion } c_i = \texttt{true}]$\\[2pt]
    {\footnotesize\bfseries $= 2.0 + 2.0 + 1.5 + 0.0 + 0.5 = \mathbf{6.0 / 8.0\text{ pts}}$}%
  };

  \draw[flow] (stage1.east |- llm.center) -- (llm.west);
  \draw[flow] (llm.east) -- (stage3.west |- llm.center);

  \draw[flow] (stage3.south) -- (stage4.north);

  \draw[dashflow, rounded corners=4pt]
    (crit.east) -- ++(11pt, 0) |- (stage4.west)
    node[pos=0.72, above, font=\tiny\bfseries, text=sumTxt] {Criterion weights $w_i$};

\end{tikzpicture}%
    }
    \caption{Overview of the multimodal AI grading pipeline for a single question response.}
    \label{fig:pipeline}
\end{figure}
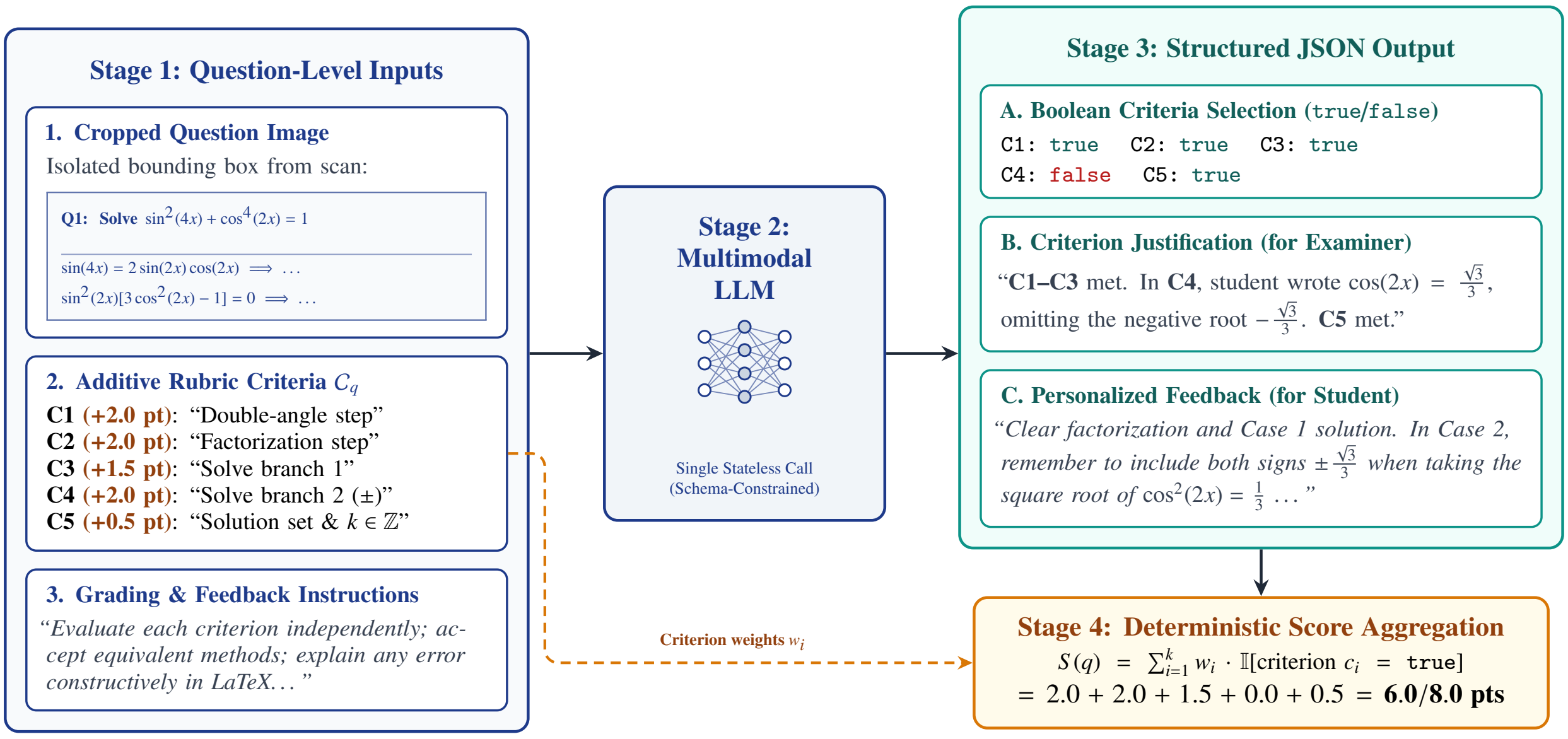

\subsection{Global Grading Pipeline}

Rather than asking an LLM to directly predict an arbitrary holistic numerical score for an entire examination script—a setting prone to hallucination, arithmetic summing errors, and opaque reasoning—the grading pipeline illustrated in Figure~\ref{fig:pipeline} decomposes assessment evaluation along two axes: \textit{spatial decomposition} (isolating individual question responses from the scanned copy) and \textit{criterion-level decomposition} (evaluating explicit rubric items while keeping numerical summation outside the LLM).

\paragraph{Question-Level Inputs (Stage~1)}
For each question $q$ with maximum score $S_{\max}(q)$, the prompt submitted to the multimodal LLM combines three components:
\begin{itemize}
    \item \textbf{Cropped Question Image:} A cropped image extracted from the student's scanned examination copy according to a rectangular bounding box defined on the blank assessment template. It contains both the printed question statement and the designated workspace where the student wrote their answer, providing the model with a focused visual input restricted to the relevant problem and handwritten work.
    \item \textbf{Additive Rubric Criteria ($\mathcal{C}_q$):} A set of $k$ explicit rubric items $\mathcal{C}_q = \{(c_1, w_1, d_1), \dots, (c_k, w_k, d_k)\}$, where each criterion $i \in \{1, \dots, k\}$ comprises a unique identifier label $c_i$ (e.g., \texttt{C1}), a positive numerical score reward $w_i > 0$ indicating the points awarded when satisfied, and a descriptive text $d_i$ specifying the mathematical property, reasoning step, or result being evaluated. Unlike manual grading grids that sometimes mix positive credit and negative penalty items within the same question, our methodology restricts criteria to strictly positive (additive) score rewards such that $\sum_{i=1}^{k} w_i = S_{\max}(q)$; a student satisfying all $k$ criteria thus obtains full marks.
    \item \textbf{Grading and Feedback Instructions:} Pedagogical and mathematical directives guiding the evaluation. The grading instructions direct the model to evaluate each criterion in $\mathcal{C}_q$ independently based on the \textit{Question Image} and encode global evaluation rules (e.g., validating criteria by mathematical equivalence when a student uses a valid alternative method, forgiving minor transcription slips in intermediate steps, and ignoring crossed-out scratch work). The feedback instructions specify how to address the student in a constructive, formative tone and explain the mathematical nature of detected errors rather than merely repeating raw rubric labels.
\end{itemize}

\begin{figure}[t]
    \centering
    \includegraphics[width=0.76\linewidth]{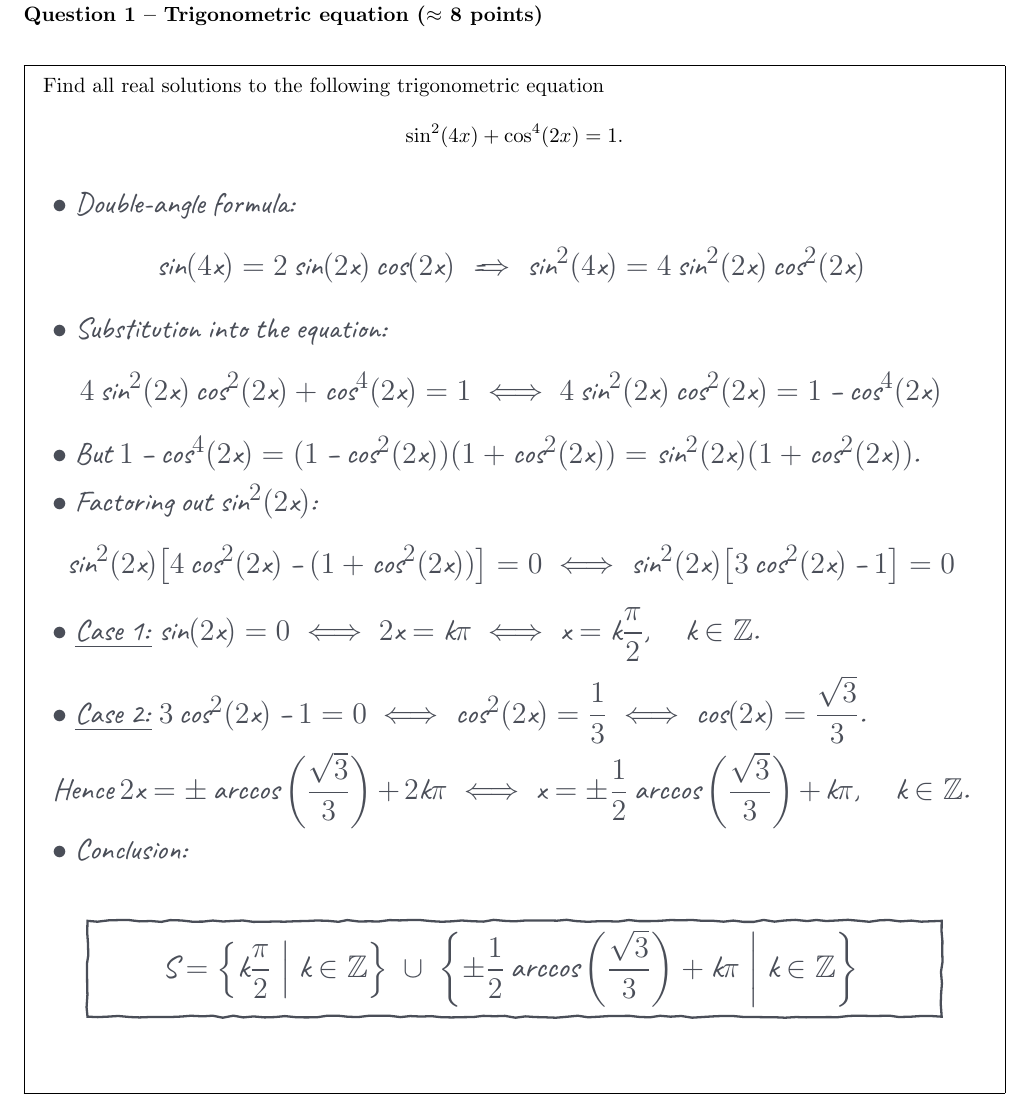}
    \caption{Illustrative cropped \textit{Question Image} for a trigonometric equation question ($S_{\max} = 8.0$ pts), containing the printed problem statement and a synthetic, LLM-generated handwritten solution created to preserve student anonymity.}
    \label{fig:example_crop}
\end{figure}

\paragraph{Stateless Multimodal Inference and Structured Outputs (Stages~2--3)}
Given the question-level prompt, the multimodal LLM is invoked in a single stateless call constrained by a JSON schema dynamically built from the question's criterion labels $\{c_1, \dots, c_k\}$.
This invocation returns three structured fields:
\begin{itemize}
    \item \textbf{Boolean Criteria Selection:} For each criterion $c_i \in \{c_1, \dots, c_k\}$, a Boolean value (\texttt{true} if the criterion is satisfied in the student's answer, or \texttt{false} otherwise).
    \item \textbf{Criterion Justification:} A detailed step-by-step evaluation intended for the examiner, explaining for each criterion $c_i$ why it is or is not satisfied by the student's handwritten derivation.
    \item \textbf{Personalized Feedback:} A formative paragraph addressed to the student, summarizing what was done correctly, pinpointing the exact nature of any mathematical errors, and offering constructive guidance for improvement.
\end{itemize}
Generating all three fields in a single stateless invocation per question avoids multi-turn prompting over the same high-resolution image, while ensuring that the evaluation of one question or student never influences another.
Moreover, because autoregressive generation conditions later tokens on earlier ones, producing the step-by-step criterion justifications prior to the feedback ensures that the student feedback is grounded in an explicit analysis of the handwriting.

\paragraph{Deterministic Score Aggregation (Stage~4)}
Crucially, the LLM never computes a numerical grade itself.
Once the Boolean criterion selections are returned by the model, the numerical grade $S(q)$ awarded to the student for question $q$ is computed deterministically outside the LLM by summing the score rewards of all selected criteria:
\[
S(q) = \sum_{i=1}^{k} w_i \cdot \mathbb{I}\big[\text{criterion } c_i \text{ is selected (\texttt{true})}\big].
\]

\subsection{Illustrative Example of Question Grading}
\label{sec:example}

To make this grading process concrete, we illustrate in Figure~\ref{fig:example_crop} how the pipeline evaluates a response to a trigonometric equation question (worth $S_{\max} = 8.0$ points), which asks the student to find all real solutions to $\sin^2(4x) + \cos^4(2x) = 1$.
To protect student privacy and avoid reproducing an authentic candidate's handwriting, the handwritten response in Figure~\ref{fig:example_crop} is a synthetic example generated for the purpose of this illustration.

\begin{figure}[t]
    \centering
    \begin{tikzpicture}[font=\small]
      \definecolor{outBody}{HTML}{F0FDFA}
      \definecolor{outHead}{HTML}{CCFBF1}
      \definecolor{outLine}{HTML}{0D9488}
      \definecolor{outTitle}{HTML}{115E59}
      \definecolor{badgePassBg}{HTML}{D1FAE5}
      \definecolor{badgePassTxt}{HTML}{115E59}
      \definecolor{badgeFailBg}{HTML}{FEE2E2}
      \definecolor{badgeFailTxt}{HTML}{B91C1C}

      \node[rectangle, rounded corners=3pt, draw=outLine!70, line width=0.5pt, fill=white,
            inner xsep=10pt, inner ysep=8pt, text width=0.92\textwidth, align=justify] (critbox) at (0,0) {%
        {\color{outTitle}\bfseries Criteria Selection \& Resulting Score}\par\vspace{4pt}
        \renewcommand{\arraystretch}{1.25}%
        \begin{tabularx}{\linewidth}{@{}l X r@{}}
          \texttt{[$\checkmark$]} \textbf{C1} & Double-angle formula $\sin^2(4x) = 4\sin^2(2x)\cos^2(2x)$ and substitution & \textbf{+2.0 / 2.0 pt} \\
          \texttt{[$\checkmark$]} \textbf{C2} & Factorization into $\sin^2(2x)[3\cos^2(2x) - 1] = 0$ & \textbf{+2.0 / 2.0 pt} \\
          \texttt{[$\checkmark$]} \textbf{C3} & Complete solution of Case 1: $\sin(2x) = 0 \iff x = k\frac{\pi}{2}$ & \textbf{+1.5 / 1.5 pt} \\
          \texttt{[$\times$]} \textbf{C4} & Complete solution of Case 2: $3\cos^2(2x) - 1 = 0$ (both signs $\pm\frac{\sqrt{3}}{3}$) & \textcolor{badgeFailTxt}{\textbf{0.0 / 2.0 pt}} \\
          \texttt{[$\checkmark$]} \textbf{C5} & Synthesis of solution set $S$ and explicit mention of $k \in \mathbb{Z}$ & \textbf{+0.5 / 0.5 pt} \\
          \midrule
          \multicolumn{2}{@{}l}{\textbf{Deterministic Total Score} $S(\text{Q1}) = 2.0 + 2.0 + 1.5 + 0.0 + 0.5$} & \textbf{6.0 / 8.0 pts}
        \end{tabularx}
      };

      \node[rectangle, rounded corners=3pt, draw=outLine!70, line width=0.5pt, fill=white,
            inner xsep=10pt, inner ysep=8pt, text width=0.92\textwidth, align=justify,
            anchor=north] (justbox) at ([yshift=-6pt]critbox.south) {%
        {\color{outTitle}\bfseries Criterion-Level Justification (for the Examiner)}\par\vspace{4pt}
        The student correctly applies the double-angle formula $\sin(4x) = 2\sin(2x)\cos(2x)$ to obtain $\sin^2(4x) = 4\sin^2(2x)\cos^2(2x)$ and substitutes it into the original equation (\textbf{C1} is met). The student rewrites $1 - \cos^4(2x)$ as $\sin^2(2x)(1 + \cos^2(2x))$ and successfully factors the equation to obtain $\sin^2(2x)[3\cos^2(2x) - 1] = 0$ (\textbf{C2} is met). Case~1 is solved correctly, yielding $x = k\frac{\pi}{2}$ with $k \in \mathbb{Z}$ (\textbf{C3} is met). However, in Case~2, the student writes $\cos^2(2x) = \frac{1}{3} \iff \cos(2x) = \frac{\sqrt{3}}{3}$, missing the negative root $\cos(2x) = -\frac{\sqrt{3}}{3}$. Consequently, the solutions generated with period $+k\pi$ miss the negative-root branches, which would have yielded $\pm\frac{1}{2}\arccos\left(\frac{\sqrt{3}}{3}\right) + k\frac{\pi}{2}$ (\textbf{C4} is not met). Finally, the synthesis of the set $S$ and the explicit condition $k \in \mathbb{Z}$ are properly presented (\textbf{C5} is met).
      };

      \node[rectangle, rounded corners=3pt, draw=outLine!70, line width=0.5pt, fill=white,
            inner xsep=10pt, inner ysep=8pt, text width=0.92\textwidth, align=justify,
            anchor=north] (fbbox) at ([yshift=-6pt]justbox.south) {%
        {\color{outTitle}\bfseries Personalized Formative Feedback (for the Student)}\par\vspace{4pt}
        \itshape The student demonstrated a clear and structured approach to transforming and factoring the trigonometric equation, and solved Case~1 perfectly. However, in Case~2, when taking the square root of $\cos^2(2x) = \frac{1}{3}$, the student omitted the negative root $\cos(2x) = -\frac{\sqrt{3}}{3}$. Taking both signs into account would have yielded the full set of solutions $x = \pm\frac{1}{2}\arccos\left(\frac{\sqrt{3}}{3}\right) + k\frac{\pi}{2}$ for $k \in \mathbb{Z}$. The student should be cautious not to omit the $\pm$ sign when solving equations of the form $X^2 = c$.
      };

      \begin{scope}[on background layer]
        \node[rectangle, rounded corners=4pt, draw=outLine, line width=0.8pt, fill=outBody,
              inner xsep=8pt, inner ysep=8pt,
              fit=(critbox)(justbox)(fbbox)] (outer) {};
      \end{scope}
    \end{tikzpicture}
    \caption{Formatted evaluation output produced by \texttt{gemini-3.8-flash} for the handwritten response in Figure~\ref{fig:example_crop}, showing the selected criteria, the examiner justification, and the personalized student feedback.}
    \label{fig:example_output}
\end{figure}

The question is decomposed into $k = 5$ strictly positive criteria whose score rewards sum to $8.0$ points:
\begin{itemize}
    \item \textbf{C1 — Double-angle formula} ($w_1 = 2.0\text{ pts}$): Correct application of the double-angle formula $\sin(4x) = 2\sin(2x)\cos(2x)$ to obtain $\sin^2(4x) = 4\sin^2(2x)\cos^2(2x)$ and substitution into the equation.
    \item \textbf{C2 — Factorization} ($w_2 = 2.0\text{ pts}$): Use of the identity $1 - \cos^4(2x) = \sin^2(2x)(1 + \cos^2(2x))$ and correct factorization yielding $\sin^2(2x)[3\cos^2(2x) - 1] = 0$ (or a valid equivalent such as a quadratic equation in $u = \cos^2(2x)$).
    \item \textbf{C3 — Solution of Case~1} ($w_3 = 1.5\text{ pts}$): Complete and exact solution of the first case $\sin(2x) = 0$, yielding $2x = k\pi$ and $x = k\frac{\pi}{2}$, with correct division of the period by $2$.
    \item \textbf{C4 — Solution of Case~2} ($w_4 = 2.0\text{ pts}$): Exact solution of the second case $3\cos^2(2x) - 1 = 0$, including consideration of both signs $\cos(2x) = \pm\frac{\sqrt{3}}{3}$ and correct division of the angle and period, yielding $x = \pm\frac{1}{2}\arccos\left(\frac{\sqrt{3}}{3}\right) + k\frac{\pi}{2}$.
    \item \textbf{C5 — Synthesis and $k \in \mathbb{Z}$} ($w_5 = 0.5\text{ pt}$): Clear synthesis of the solution set $S$ and explicit mention that the periodicity constant belongs to $\mathbb{Z}$ ($k \in \mathbb{Z}$).
\end{itemize}
Accompanying these criteria, the \textit{Grading Instructions} direct the model to evaluate each criterion independently, to accept mathematically equivalent resolution paths, and to invalidate the corresponding solving criterion (\textbf{C3} or \textbf{C4}) if the student forgets to divide the period by $2$ when isolating $x$.

\begin{table*}[t]
  \centering
  \caption{Summary of the four exams and of the grades given by the human graders on Gradescope.
    \emph{\# Questions} counts main questions only (not subquestions).
    \emph{\# Human Graders} is the number of graders listed on Gradescope.
    \emph{Scale} is the range of points achievable on the exam.
    Std is the sample standard deviation.
    \emph{Min} and \emph{Max} are the lowest and highest total grades observed on the students' copies.
    \emph{Pass} is the share of copies scoring at least half of the maximum.}
  \label{tab:exam-summary}
  \small
  \setlength{\tabcolsep}{5pt}
  \begin{tabular}{l cccr rrrrrr}
    \toprule
    & \multicolumn{4}{c}{Dataset} & \multicolumn{6}{c}{Human grades (pts)} \\
    \cmidrule(lr){2-5} \cmidrule(lr){6-11}
    Exam & Copies & \# Human Graders & \# Questions & Scale & Mean & Std & Median & Min & Max & Pass \\
    \midrule
    Geometry     & 435 & 4 & 4 & 0--40 & 16.18 &  8.56 & 16.50 & 0 & 37.5 & 37.5\% \\
    Trigonometry & 442 & 4 & 3 & 0--20 & 10.20 &  4.77 & 11.00 & 0 & 20   & 65.2\% \\
    Algebra      & 438 & 4 & 3 & 0--20 &  8.70 &  4.63 &  8.50 & 0 & 20   & 39.5\% \\
    Analysis     & 437 & 7 & 4 & 0--50 & 20.34 & 10.86 & 20.00 & 0 & 48   & 34.1\% \\
    \bottomrule
  \end{tabular}
\end{table*}

Given the cropped image in Figure~\ref{fig:example_crop} and the rubric above, \texttt{gemini-3.8-flash} produces the evaluation rendered in Figure~\ref{fig:example_output} (the corresponding JSON schema and raw payload are provided in \ref{app:json}).
The model validates \textbf{C1}, \textbf{C2}, \textbf{C3}, and \textbf{C5} ($6.0/8.0$ points) while withholding \textbf{C4} and pinpointing the omitted negative root $-\frac{\sqrt{3}}{3}$ in both the examiner justification and the student feedback.

\section{Benchmarks and Multimodal LLMs}
\label{sec:setup}

To assess how accurately multimodal LLMs grade authentic handwritten assessments, we evaluate the proposed grading pipeline across four real-world university entrance examinations and compare representative frontier proprietary and open-weights multimodal models.

\subsection{Benchmarks}
Our benchmark dataset comes from the June 2026 session of the mandatory mathematics entrance examination for admission to civil engineering programs in Wallonia (Belgium), administered over two days at UCLouvain.
Every year, between 400 and 500 prospective engineering students sit this examination, which is divided into four independent written tests: Geometry, Trigonometry, Algebra, and Analysis (Calculus).
Each test is composed of three to four main open-ended questions, where expected answers range from a single numerical value or symbolic expression to complex multi-step mathematical derivations, geometric constructions, or tables.
All scripts are handwritten by students on structured paper templates and subsequently scanned, yielding between 435 and 442 graded copies per subject (1,752 copies in total).

For several years, this entrance examination has been graded using Gradescope.
For each of the four subjects, two university professors design the examination questions and construct the corresponding analytic grading rubrics; however, due to the large number of copies and the strict turnaround deadlines required to publish admission results, a team of 4 to 7 human graders (professors and teaching assistants) is mobilized per subject to carry out the manual grading on Gradescope.
These official Gradescope evaluations serve as the human ground-truth baseline against which we evaluate automated AI grading.
Complete details on the four exams, their grading scales, and the human score distributions are reported in Table~\ref{tab:exam-summary}.

To configure the multimodal AI grading pipeline for these four examinations, we preserved the exact point weighting of every question from the official exam.
We started directly from the Gradescope rubrics created by the instructors for each question and used an LLM to make the criterion descriptions more explicit and self-contained (as human Gradescope items are often written as brief reminders for fellow instructors).
Whereas some instructors combined positive (additive) and negative (deductive penalty) rubric items within the same question on Gradescope, we standardized all AI rubrics to use strictly positive criteria, ensuring that the sum of criterion weights equals the question's maximum score and that a student validating every criterion receives full marks.
Finally, prior to grading the full cohort, we calibrated the wording of these criteria on at most 4 to 5 copies per question to verify that a poor copy was consistently graded as poor, an average copy remained average, and a strong copy obtained a high score.
All experiments were conducted using Gradaly\footnote{\url{https://gradaly.io}}, a desktop application developed by the first author that implements the pipeline of Section~\ref{sec:approach} and provides an interactive interface to define question cropping regions on the examination template, edit and calibrate the additive rubric criteria, and execute the multimodal LLM evaluations.

\subsection{Evaluated Multimodal LLMs}

To evaluate both the overall alignment of AI grading with human examiners (RQ1) and the competitiveness of open-weights models against proprietary frontier models (RQ2) across the entire 1,752-copy benchmark, we evaluate two representative multimodal Vision-Language Models:
\begin{itemize}
    \item \textbf{Google Gemini 3.8 Flash (\texttt{gemini-3.8-flash}):} A proprietary multimodal model developed by Google DeepMind belonging to the lightweight ``Flash'' tier. It is engineered for high-throughput, low-latency multimodal reasoning, combining native high-resolution vision encoding—well suited for reading dense handwritten mathematical notation, crossed-out derivations, and geometric sketches—with structured output generation and internal reasoning tokens. Notably, while \texttt{gemini-3.8-flash} is already a strong model, it is not the largest model in the Gemini family: Google offers even more capable flagship models (such as the \texttt{Pro} tier), but \texttt{gemini-3.8-flash} provides a highly attractive balance between reasoning quality and inference speed (a few seconds per copy) for large-scale cohort grading.
    \item \textbf{GLM 5.3 Flash (\texttt{glm-5.3-flash}):} An open-weights multimodal Vision-Language Model from the General Language Model (GLM) family developed by Zhipu AI (THUDM). Designed as a compact, fast-inference multimodal model capable of joint visual-textual reasoning over document images and diagrams, \texttt{glm-5.3-flash} represents a compelling open-weights alternative that can be deployed either via third-party inference providers or on local GPU infrastructure.
\end{itemize}

Both models are evaluated in a zero-shot setting without task-specific fine-tuning.
Because the entrance examinations were held in June 2026, none of the models could have encountered these exam problems or solutions during pre-training, ruling out data contamination.

\section{Empirical Evaluation}
\label{sec:results}

In this section, we experimentally address our research questions by evaluating the multimodal grading pipeline across the four mathematics entrance examinations (1,752 copies in total).
We first examine the alignment of AI grades with human Gradescope evaluations (RQ1) and compare the frontier \texttt{gemini-3.8-flash} model against the open-weights \texttt{glm-5.3-flash} model (RQ2).
We then assess whether the generated feedback is specific to each student's answer (RQ3), analyze the usage of the grading criteria across each cohort (RQ4), and investigate run-to-run grading stability (RQ5).

\subsection{Alignment with Human Grading and Comparison of Frontier vs.\ Open-Weights Models (RQ1 \& RQ2)}

\begin{table*}[t]
  \centering
  \caption{Comparison of automated grading performance between \texttt{glm-5.3-flash} (open-weights) and \texttt{gemini-3.8-flash} (proprietary frontier Flash tier) against the human Gradescope baseline across all four mathematics entrance examinations ($N = 1{,}752$ copies total).
  \emph{Bias} is the mean signed difference ($\text{AI} - \text{Human}$); \emph{MAE} is the Mean Absolute Error (with percentage of total exam scale in parentheses); $r$, $\rho$, and $\text{CCC}$ denote Pearson linear correlation, Spearman rank correlation, and Lin's Concordance Correlation Coefficient, respectively; \emph{Pass Agr.} is the classification agreement at the $50\%$ pass threshold.}
  \label{tab:main_comparison}
  \footnotesize
  \setlength{\tabcolsep}{4.5pt}
  \begin{adjustbox}{max width=\textwidth}
  \begin{tabular}{ll rrrrrrrr}
    \toprule
    Exam (Scale, $N$) & Model & Mean $\pm$ Std & Bias (pts / \%) & MAE (pts / \%) & RMSE & Pearson $r$ & Spearman $\rho$ & Lin's CCC & Pass Agr. \\
    \midrule
    \textbf{Geometry} & Human Baseline & $16.18 \pm 8.56$ & --- & --- & --- & --- & --- & --- & --- \\
    (0--40 pts, $N=435$) & \texttt{glm-5.3-flash} & $15.90 \pm 8.67$ & $\mathbf{-0.28}$ ($-0.7\%$) & $2.39$ ($6.0\%$) & $3.24$ & $0.930$ & $0.929$ & $0.931$ & $89.7\%$ \\
     & \texttt{gemini-3.8-flash} & $15.09 \pm 8.54$ & $-1.10$ ($-2.7\%$) & $\mathbf{1.89}$ ($\mathbf{4.7\%}$) & $\mathbf{2.49}$ & $\mathbf{0.966}$ & $\mathbf{0.966}$ & $\mathbf{0.960}$ & $\mathbf{92.0\%}$ \\
    \midrule
    \textbf{Analysis} & Human Baseline & $20.34 \pm 10.86$ & --- & --- & --- & --- & --- & --- & --- \\
    (0--50 pts, $N=437$) & \texttt{glm-5.3-flash} & $25.19 \pm 11.68$ & $+4.85$ ($+9.7\%$) & $5.92$ ($11.8\%$) & $7.64$ & $0.865$ & $0.858$ & $0.791$ & $77.1\%$ \\
     & \texttt{gemini-3.8-flash} & $20.87 \pm 11.49$ & $\mathbf{+0.53}$ ($\mathbf{+1.1\%}$) & $\mathbf{2.78}$ ($\mathbf{5.6\%}$) & $\mathbf{3.82}$ & $\mathbf{0.944}$ & $\mathbf{0.940}$ & $\mathbf{0.944}$ & $\mathbf{89.0\%}$ \\
    \midrule
    \textbf{Algebra} & Human Baseline & $8.70 \pm 4.63$ & --- & --- & --- & --- & --- & --- & --- \\
    (0--20 pts, $N=438$) & \texttt{glm-5.3-flash} & $9.91 \pm 4.44$ & $+1.20$ ($+6.0\%$) & $1.89$ ($9.5\%$) & $2.45$ & $0.889$ & $0.883$ & $0.860$ & $82.0\%$ \\
     & \texttt{gemini-3.8-flash} & $8.62 \pm 4.53$ & $\mathbf{-0.09}$ ($\mathbf{-0.4\%}$) & $\mathbf{1.19}$ ($\mathbf{6.0\%}$) & $\mathbf{1.58}$ & $\mathbf{0.940}$ & $\mathbf{0.934}$ & $\mathbf{0.942}$ & $\mathbf{88.6\%}$ \\
    \midrule
    \textbf{Trigonometry} & Human Baseline & $10.20 \pm 4.77$ & --- & --- & --- & --- & --- & --- & --- \\
    (0--20 pts, $N=442$) & \texttt{glm-5.3-flash} & $11.25 \pm 4.61$ & $+1.05$ ($+5.2\%$) & $2.56$ ($12.8\%$) & $3.35$ & $0.769$ & $0.711$ & $0.752$ & $81.9\%$ \\
     & \texttt{gemini-3.8-flash} & $9.53 \pm 4.95$ & $\mathbf{-0.67}$ ($\mathbf{-3.4\%}$) & $\mathbf{1.42}$ ($\mathbf{7.1\%}$) & $\mathbf{2.07}$ & $\mathbf{0.919}$ & $\mathbf{0.890}$ & $\mathbf{0.912}$ & $\mathbf{85.3\%}$ \\
    \bottomrule
  \end{tabular}
  \end{adjustbox}
\end{table*}

\begin{figure}[t]
    \centering
    \includegraphics[width=\linewidth]{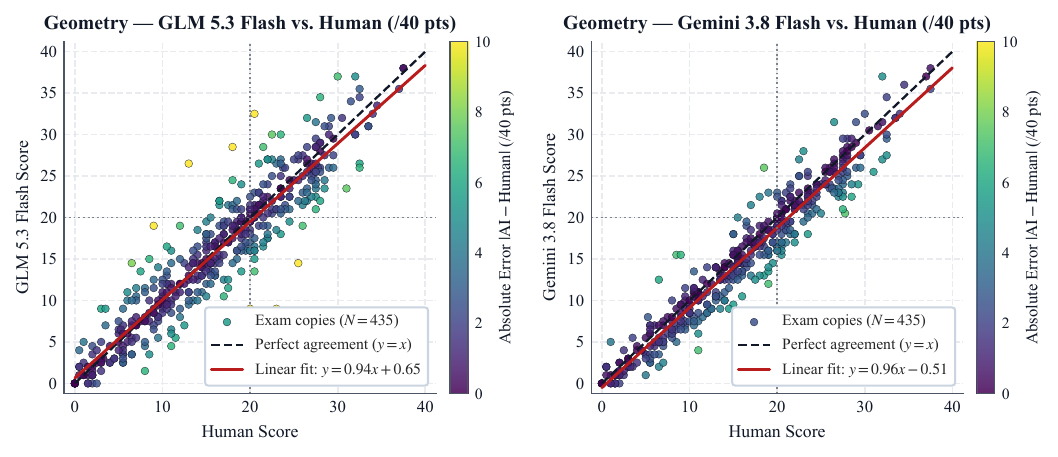}
    \caption{Total score scatter plot and linear regression on the \textbf{Geometry} examination ($N=435$, scale $/40$ pts) for \texttt{glm-5.3-flash} (left) and \texttt{gemini-3.8-flash} (right) against human Gradescope scores.}
    \label{fig:scatter_geom}
\end{figure}

\begin{figure}[t]
    \centering
    \includegraphics[width=\linewidth]{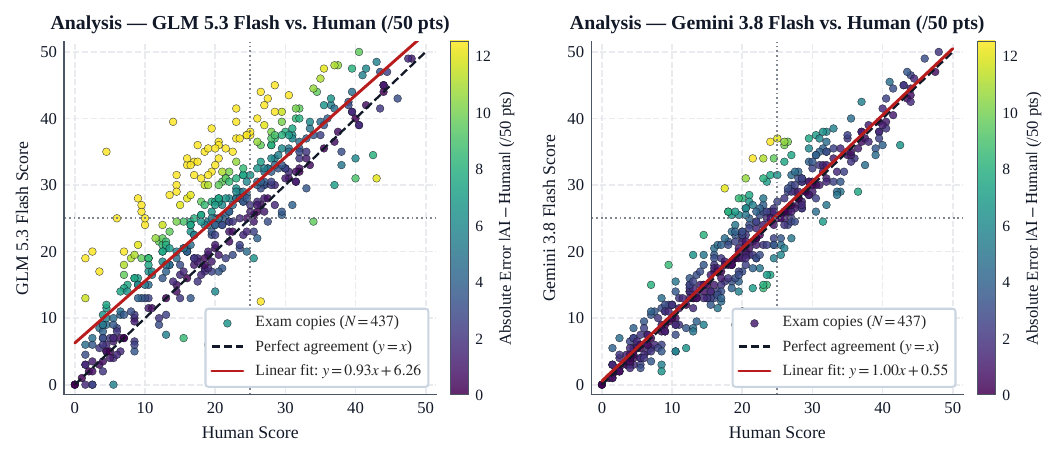}
    \caption{Total score scatter plot and linear regression on the \textbf{Analysis (Calculus)} examination ($N=437$, scale $/50$ pts) for \texttt{glm-5.3-flash} (left) and \texttt{gemini-3.8-flash} (right) against human Gradescope scores. Scatter plots for Algebra and Trigonometry are reported in \ref{app:criterion_usage}.}
    \label{fig:scatter_ana}
\end{figure}

Table~\ref{tab:main_comparison} summarizes the inter-rater agreement metrics across all four examinations, while Figures~\ref{fig:scatter_geom} and \ref{fig:scatter_ana} display representative copy-level scatter plots for Geometry and Analysis comparing total scores awarded by the human graders (horizontal axis) against those awarded by \texttt{glm-5.3-flash} (left panel) and \texttt{gemini-3.8-flash} (right panel); the corresponding scatter plots for Algebra and Trigonometry are provided in \ref{app:criterion_usage}.

\paragraph{High Alignment of \texttt{gemini-3.8-flash} with Human Graders (RQ1)}
As reported in Table~\ref{tab:main_comparison} and illustrated in the right-hand panels of Figures~\ref{fig:scatter_geom} and \ref{fig:scatter_ana}, \texttt{gemini-3.8-flash} achieves a remarkably tight alignment with human Gradescope grading across all four mathematical disciplines.
Pearson linear correlation reaches $r = 0.966$ ($R^2 = 93.3\%$) on Geometry, $r = 0.944$ ($R^2 = 89.1\%$) on Analysis, $r = 0.940$ ($R^2 = 88.4\%$) on Algebra, and $r = 0.919$ ($R^2 = 84.5\%$) on Trigonometry.
More importantly, because high correlation alone does not guarantee calibration if a grader exhibits systematic shift or scale compression, Lin's Concordance Correlation Coefficient ($\text{CCC}$) confirms strong absolute agreement alongside correlation, ranging from $\text{CCC} = 0.912$ (Trigonometry) to $\text{CCC} = 0.960$ (Geometry).

Across all four subjects, the linear regression fit (solid red line) tracks the perfect-agreement diagonal $y = x$ (dashed black line) very closely, with slopes between $0.92$ and $1.00$ ($y = 1.00x + 0.55$ in Analysis and $y = 0.96x - 0.51$ in Geometry).
Systematic bias is minimal: the mean difference between \texttt{gemini-3.8-flash} and the human graders is only $-0.09$ pts ($-0.4\%$ of the scale) in Algebra, $+0.53$ pts ($+1.1\%$) in Analysis, $-1.10$ pts ($-2.7\%$) in Geometry, and $-0.67$ pts ($-3.4\%$) in Trigonometry.
The Mean Absolute Error (MAE) remains consistently between $4.7\%$ and $7.1\%$ of the total exam scale ($1.89/40$ pts in Geometry, $2.78/50$ pts in Analysis, $1.19/20$ pts in Algebra, and $1.42/20$ pts in Trigonometry).
At the standard $50\%$ pass/fail threshold, \texttt{gemini-3.8-flash} agrees with the human pass/fail decision on $85.3\%$ to $92.0\%$ of scripts, with nearly all disagreements occurring on borderline copies whose human scores lie within $\pm 1.5$ points of the threshold.
It is worth emphasizing that this level of accuracy is achieved using a fast, lightweight ``Flash'' model without any task-specific fine-tuning; even more capable frontier reasoning models in the Gemini lineup could further reduce residual errors on complex proofs, albeit at higher latency.

Crucially, a manual audit of the few outlier copies showing a pronounced discrepancy between the human Gradescope score and the \texttt{gemini-3.8-flash} evaluation revealed that, on several occasions, the AI assessment was in fact correct and exposed clerical errors in the human reference grading.
For instance, when grading hundreds of scripts sequentially in the Gradescope interface, human examiners occasionally clicked the wrong rubric checkbox when two contrasting options were positioned immediately next to each other—such as accidentally selecting the ``$100\%$ correct answer'' item (awarding full credit) instead of the adjacent ``question not answered'' item (awarding zero points) on a blank response.
Whereas clerical errors that penalized students were generally identified and rectified when candidates reviewed their graded copies, errors made in a student's favor were understandably never reported by the beneficiaries and thus persisted in the human ground-truth dataset.
This finding highlights that a non-negligible share of the residual gap between the LLM and the human benchmark stems from human occasion noise and interface misclicks rather than model failure, and demonstrates the utility of the pipeline as a systematic quality-assurance mechanism.

\paragraph{Comparison with Open-Weights \texttt{glm-5.3-flash} (RQ2)}
Comparing the two models in Table~\ref{tab:main_comparison} and the left and right panels of Figures~\ref{fig:scatter_geom} and \ref{fig:scatter_ana} reveals both the promise and the current limitations of compact open-weights multimodal models relative to frontier proprietary models:
\begin{itemize}
    \item \textbf{Strong performance on structured geometric and algebraic tasks:} On Geometry (Figure~\ref{fig:scatter_geom}, left), the open-weights \texttt{glm-5.3-flash} model achieves a high correlation with human graders ($r = 0.930$, $\text{CCC} = 0.931$) and an almost negligible mean bias of $-0.28$ pts ($-0.7\%$), reaching $89.7\%$ pass/fail agreement. On Algebra (Table~\ref{tab:main_comparison}), it similarly attains a solid correlation of $r = 0.889$ ($\text{CCC} = 0.860$, MAE $= 1.89/20$ pts). This demonstrates that modern open-weights vision-language models can already capture the overall structure of handwritten mathematical solutions.
    \item \textbf{Leniency bias and higher dispersion on multi-step calculus and trigonometry:} Despite these encouraging results, \texttt{gemini-3.8-flash} consistently outperforms \texttt{glm-5.3-flash} across every exam, reducing the MAE by $21\%$ on Geometry ($1.89$ vs.\ $2.39$ pts), $37\%$ on Algebra ($1.19$ vs.\ $1.89$ pts), $45\%$ on Trigonometry ($1.42$ vs.\ $2.56$ pts), and $53\%$ on Analysis ($2.78$ vs.\ $5.92$ pts).
    In Analysis (Figure~\ref{fig:scatter_ana}, left), \texttt{glm-5.3-flash} exhibits a pronounced positive leniency bias of $+4.85$ pts ($+9.7\%$ of the scale, regression intercept $+6.26$), visible as a cluster of yellow/green points shifted above the $y = x$ diagonal for copies scoring between $10$ and $30/50$. Inspection of individual justifications indicates that when students write lengthy multi-step calculus derivations containing subtle algebraic or limit-evaluation errors, \texttt{glm-5.3-flash} is more prone to validating intermediate criteria based on surface formula resemblance, whereas \texttt{gemini-3.8-flash} more reliably verifies the mathematical validity of each transition.
    \item \textbf{Sensitivity to equivalent trigonometric forms:} The largest correlation gap appears in Trigonometry (Table~\ref{tab:main_comparison}), where \texttt{glm-5.3-flash} drops to $r = 0.769$ ($R^2 = 59.2\%$) compared to $r = 0.919$ ($R^2 = 84.5\%$) for \texttt{gemini-3.8-flash}. Trigonometric problems frequently admit multiple valid transformation paths and require distinguishing legitimate identity substitutions from sign or periodicity errors in dense handwriting—a high-precision visual and symbolic task where the frontier model maintains a decisive advantage.
\end{itemize}

\begin{figure}[t]
    \centering
    \includegraphics[width=\linewidth]{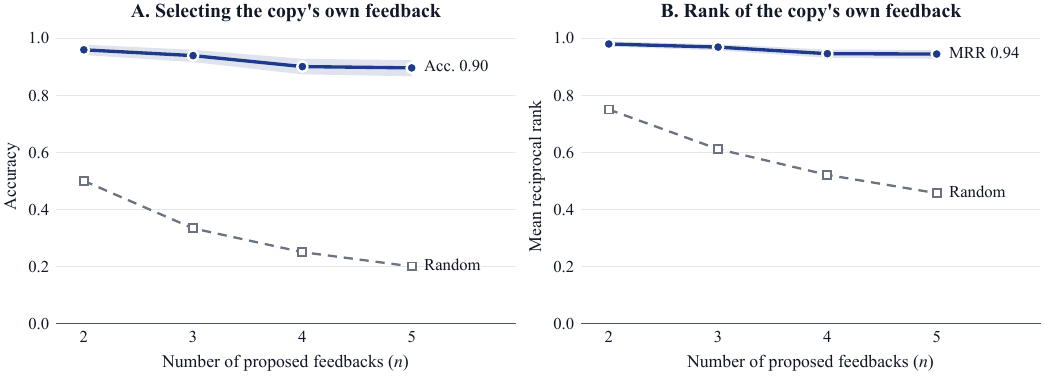}
    \caption{Identification of a copy's own feedback among $n$ candidates on Question~1 of the \textbf{Trigonometry} examination ($N=442$ copies, judge: \texttt{gemini-3.8-flash}). (A)~Top-1 accuracy: share of copies for which the judge ranks the copy's own feedback first. (B)~Mean reciprocal rank of the copy's own feedback. Shaded bands are $95\%$ percentile bootstrap confidence intervals over copies ($2{,}000$ resamples); dashed lines indicate the random-chance level.}
    \label{fig:feedback_specificity}
\end{figure}

\begin{figure}[t]
    \centering
    \includegraphics[width=\linewidth]{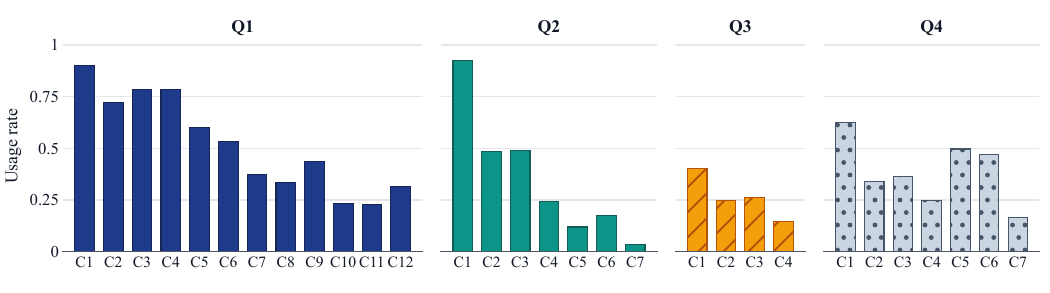}
    \caption{Usage rate of each grading criterion, per question, for the \textbf{Analysis} examination ($N=437$ copies, \texttt{gemini-3.8-flash}). The corresponding distributions for Algebra, Trigonometry, and Geometry are reported in \ref{app:criterion_usage}.}
    \label{fig:crit_ana}
\end{figure}

\subsection{Specificity of the Generated Feedback (RQ3)}
\label{sec:feedback}

Beyond the numerical grade, the pipeline returns a personalized feedback paragraph for every answer (Section~\ref{sec:approach}).
Such feedback is only formative if it engages with what the individual student actually wrote.
A known risk of LLM-generated feedback is that it collapses into generic, template-like text---for instance, a paraphrase of the grading criteria---that could be attached interchangeably to many copies.
Measuring the pedagogical usefulness of feedback ultimately requires judgments from students and instructors, which we leave for future work (Section~\ref{sec:limitations}).
Here, we evaluate a necessary condition for usefulness that can be measured at scale without human annotation: \textit{specificity}, i.e., whether the feedback written for a copy can be told apart from the feedback written for other copies of the same question, given only the student's answer.

We frame specificity as an identification task on Question~1 of the Trigonometry examination.
We use the feedback generated by \texttt{gemini-3.8-flash} for all $442$ copies during the grading run analyzed for RQ1.
For each copy, a multimodal LLM judge, \texttt{gemini-3.8-flash}, receives the cropped image of the student's answer together with $n$ candidate feedbacks: the copy's own feedback and $n-1$ distractors drawn at random, from the feedbacks of the $441$ other copies.
The candidates are shuffled and labelled with letters, and the judge is asked to rank all $n$ feedbacks from the most to the least appropriate for this answer, based on what the student actually wrote (methods, computations, errors, missing or present steps, and final results).
Importantly, although LLM evaluators can exhibit self-preference bias when comparing their own outputs against texts authored by humans or other models \citep{zheng2023judging, panickssery2024llm}, this bias is neutralized by construction in our within-model identification setup: because both the target feedback and all $n-1$ distractors were generated by the exact same model (\texttt{gemini-3.8-flash}) under identical prompt instructions, superficial stylistic cues cannot distinguish the correct candidate, forcing the judge to rely on mathematical and semantic correspondence with the handwritten image.
We repeat the task for $n \in \{2, 3, 4, 5\}$, drawing new distractors for each $n$, which yields $4 \times 442 = 1{,}768$ trials, all of which returned a valid ranking.

We report two metrics: the top-1 \textit{accuracy}, i.e., the share of copies for which the judge ranks the copy's own feedback first, and the \textit{mean reciprocal rank} (MRR) of the copy's own feedback, which also credits near misses.
A judge ranking the candidates at random achieves an expected accuracy of $1/n$ and an expected MRR of $\frac{1}{n}\sum_{j=1}^{n} \frac{1}{j}$.
The $95\%$ confidence intervals (CIs) are obtained with a percentile bootstrap over copies ($2{,}000$ resamples).

Figure~\ref{fig:feedback_specificity} shows that the judge recovers the copy's own feedback far above chance for every $n$.
With two candidates, the correct feedback is selected for $95.9\%$ of the copies ($95\%$ CI $[94.1, 97.7]$) and, with five candidates, accuracy remains at $89.6\%$ ($95\%$ CI $[86.7, 92.3]$).
The MRR follows the same trend, decreasing only from $0.980$ ($95\%$ CI $[0.971, 0.989]$) at $n = 2$ to $0.944$ ($95\%$ CI $[0.928, 0.959]$) at $n = 5$, against chance levels of $0.750$ and $0.457$.
When the judge misses, it usually ranks the correct feedback second ($38$ of the $46$ errors at $n = 5$), so that the copy's own feedback is among the top two candidates for $98.2\%$ of the copies even with five candidates.

These results indicate that the generated feedback is specific to each student's answer rather than a generic restatement of the grading criteria: it reflects the particular derivation, errors, and omissions of each copy, to the point that it can be matched back to that copy with high reliability.
The residual confusions are concentrated on near-identical, nearly complete solutions, for which similar feedback is expected and pedagogically harmless.

\begin{figure}[t]
    \centering
    \includegraphics[width=\linewidth]{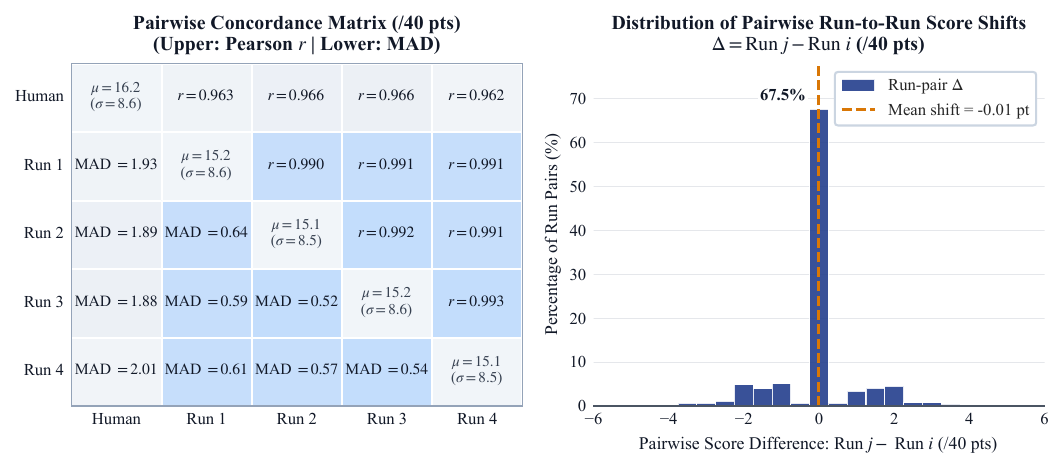}
    \caption{Pairwise run-to-run concordance across 4 independent grading runs of the Geometry examination ($N=435$, $/40$ pts) with \texttt{gemini-3.8-flash}: matrix of pairwise Pearson correlations (upper triangle) and Mean Absolute Differences (MAD, lower triangle) between the human baseline and the 4 runs, with each grader's mean and standard deviation on the diagonal (left), and distribution of pairwise score shifts $\Delta = \text{Run } j - \text{Run } i$ across $2{,}610$ run pairs (right).}
    \label{fig:var_concordance}
\end{figure}

\begin{figure}[t]
    \centering
    \includegraphics[width=\linewidth]{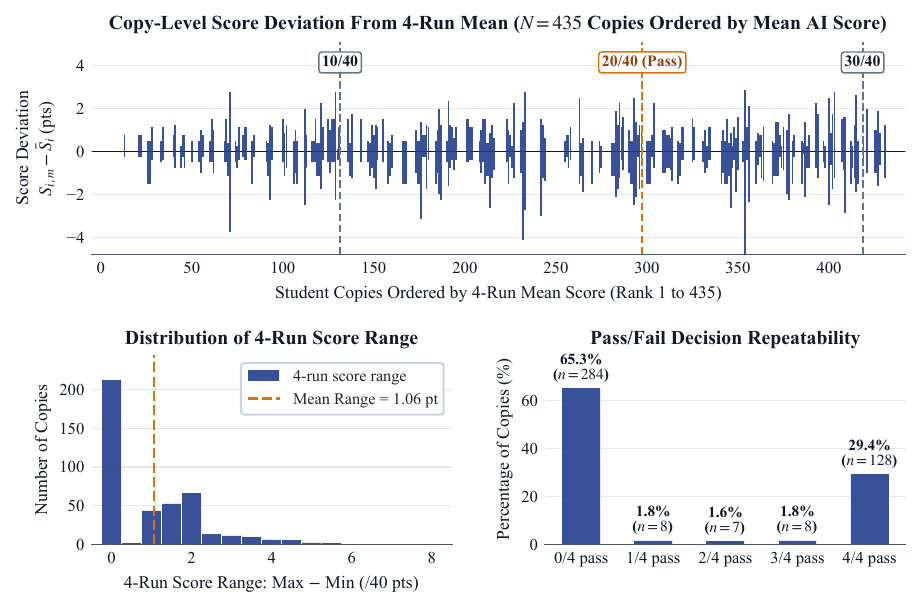}
    \caption{Copy-level score spread across the 4 independent grading runs on Geometry ($N=435$): individual student min--max score deviations from each copy's 4-run mean score ordered by mean score with vertical score-threshold markers at $10/40$, $20/40$ (pass), and $30/40$ (top), distribution of the 4-run score range $\max - \min$ per copy (bottom left), and repeatability of the pass/fail classification ($\ge 20/40$ pts) across the 4 runs (bottom right).}
    \label{fig:var_spread}
\end{figure}

\subsection{Criterion Usage (RQ4)}

To address our fourth research question (RQ4)—whether the multimodal LLM actively and meaningfully uses all defined criteria across the cohort rather than collapsing onto a subset of coarse items—we analyze the \textit{usage rate} of every criterion for \texttt{gemini-3.8-flash}, i.e., the share of copies in which the criterion is validated.
Across all $94$ criteria of the four examinations, every criterion is actively used, with validation rates ranging from $3.4\%$ to $92.7\%$; no criterion is ever satisfied on all copies or on none (the criterion usage distributions for Algebra, Trigonometry, and Geometry are provided in \ref{app:criterion_usage}).

Figure~\ref{fig:crit_ana} illustrates this behavior on the Analysis examination ($30$ criteria across four questions).
Even the rarest criterion—the final result of the second journey in Question~2 (\texttt{C7})—is validated on $15$ of the $437$ copies ($3.4\%$), whereas the two most frequently validated criteria—identifying the function's domain in Question~1 (\texttt{C1}, $90.2\%$) and stating the basic physical and financial relations in Question~2 (\texttt{C1}, $92.7\%$)—are still missed by $7\%$ to $10\%$ of the cohort.
No criterion is therefore redundant, tailored to an isolated script, or granted indiscriminately.

Moreover, the usage profile within each question reflects its pedagogical structure.
In sequential multi-step problems where each step builds on the previous ones, validation rates decrease steadily along the resolution path so that later criteria discriminate among the strongest copies: in Figure~\ref{fig:crit_ana}, usage drops from $90.2\%$ for the domain (\texttt{C1}) down to $23\%$--$31\%$ for the final asymptotic and graph-sketching steps (\texttt{C10}--\texttt{C12}) in Question~1 (function study), and from $92.7\%$ for the initial model setup (\texttt{C1}) down to $17.4\%$ and $3.4\%$ for setting up and solving the system of the second journey (\texttt{C6}--\texttt{C7}) in Question~2 (maritime transport cost optimization).
In contrast, when a question comprises more independent parts—such as Question~4 in Figure~\ref{fig:crit_ana}, where criteria \texttt{C5} and \texttt{C6} evaluate a separate sub-problem after a harder intermediate step (\texttt{C4}), or in exercises composed of independent short items—usage rates reflect the intrinsic difficulty of each sub-question rather than its position.

\subsection{Run-to-Run Grading Stability (RQ5)}
\label{sec:variability}

Because LLM inference involves stochastic sampling and high-dimensional visual-textual reasoning, a critical requirement for high-stakes educational assessment is \textit{test-retest reliability} (RQ5): if the exact same examination cohort is graded multiple times independently by the same model, do individual students receive consistent grades, or does stochasticity introduce unacceptable occasion noise?
To answer this question, we executed $M = 4$ completely independent grading runs of the entire Geometry examination ($N = 435$ copies, $4$ questions, $28$ rubric criteria, scale $/40$ points) using \texttt{gemini-3.8-flash}.

\paragraph{Pairwise Run-to-Run Concordance and Macro-Level Stability}
Figure~\ref{fig:var_concordance} summarizes the consistency across all $\binom{4}{2} = 6$ pairs of independent runs ($6 \times 435 = 2{,}610$ copy-level comparisons).
At the cohort level, the four runs exhibit virtually identical global alignment with the human baseline: the AI mean score across the 435 copies fluctuates by less than $0.1$ point across runs ($\mu = 15.16$, $15.09$, $15.18$, and $15.10/40$ pts for Runs~1 to~4), Pearson correlation with human graders remains tightly within $r \in [0.9617, 0.9663]$, and MAE versus human graders stays within $[1.88, 2.01]$ pts.
Averaging the scores across the 4 runs yields a slight ensemble gain against the human baseline ($r = 0.9673$, $\text{MAE} = 1.87$ pts).

Comparing the AI runs directly against one another reveals exceptionally high test-retest reliability:
\begin{itemize}
    \item The single-run Intraclass Correlation Coefficient (two-way random effects, absolute agreement; \citealp{shrout1979intraclass}) is $\text{ICC}(2,1) = 0.9912$ ($\text{ICC}(2,4) = 0.9978$ for the 4-run average), well above the $0.90$ threshold for excellent reliability \citep{koo2016guideline}, while the mean pairwise Pearson correlation between any two runs is $r = 0.9913$ (ranging from $0.9902$ to $0.9925$).
    \item As shown in the lower triangle of the matrix in Figure~\ref{fig:var_concordance} (left), the mean pairwise Mean Absolute Difference (MAD) between two independent runs is only $\mathbf{0.578\text{ pt}}$ on the 40-point scale ($1.4\%$ of the maximum grade)—more than three times smaller than the discrepancy between AI and human graders ($\text{MAE} \approx 1.9$ pts).
    \item Across all $2{,}610$ pairwise copy comparisons (Figure~\ref{fig:var_concordance}, right), the score shift $\Delta = \text{Run } j - \text{Run } i$ is symmetrically centered at zero (mean shift $= -0.01\text{ pt}$, $\text{SD} = 1.14\text{ pt}$): in $\mathbf{67.5\%}$ of run pairs, the student receives the \textit{exact same total score} ($\Delta = 0$) out of 40 points, and in $\mathbf{77.2\%}$ of pairs the difference is at most $1.0\text{ pt}$ ($2.5\%$ of the scale).
\end{itemize}

\paragraph{Copy-Level Score Range and Pass/Fail Repeatability}
To examine how run-to-run variability affects individual students across all four runs simultaneously, Figure~\ref{fig:var_spread} plots the minimum-to-maximum score range $\text{Range}_i = \max_{m \in \{1..4\}} S_{i,m} - \min_{m \in \{1..4\}} S_{i,m}$ for each of the 435 copies.
Remarkably, $\mathbf{49.0\%}$ of students ($213/435$) achieve \textit{exact 4/4 unanimity} ($\text{Range} = 0\text{ pt}$), receiving the exact same total score across all four independent executions.
The mean 4-run range across the entire cohort is only $1.06\text{ pts}$ (median $= 1.0\text{ pt}$), with a mean within-copy standard deviation of $\sigma = 0.52\text{ pt}$.
Overall, $59.5\%$ of copies vary by at most $1.0\text{ pt}$ across the four runs, $87.1\%$ ($379/435$) vary by at most $2.0\text{ pts}$ ($5\%$ of the exam scale), and $93.1\%$ ($405/435$) vary by at most $3.0\text{ pts}$.

From a decision-making perspective, the bottom-right panel of Figure~\ref{fig:var_spread} evaluates the stability of binary pass/fail outcomes at the official $20/40$ threshold.
Across all four grading runs, $\mathbf{94.7\%}$ of candidates ($412/435$) receive a unanimous $4/4$ classification ($65.3\%$ unanimous fail and $29.4\%$ unanimous pass).
Only $23$ copies ($5.3\%$) experience a split pass/fail outcome across the four runs, and their mean AI score is $19.88/40$—confirming that pass/fail flips are strictly confined to borderline scripts sitting immediately on the $20.0$-point cutoff.

\paragraph{Criterion-Level Decision Repeatability}
Finally, we examine the microscopic source of run-to-run variance across all $28 \text{ criteria} \times 435 \text{ copies} = 12{,}180$ Boolean criterion evaluations per run on the Geometry examination (the per-criterion breakdown is provided in \ref{app:criterion_usage}).
At the cohort level, the validation rates of all 28 criteria are virtually indistinguishable across Runs~1 to~4.
At the individual copy-criterion level, the mean pairwise agreement across the 6 run pairs is $\mathbf{98.39\%}$, and $\mathbf{97.07\%}$ of all $12{,}180$ copy-criterion pairs achieve unanimous $4/4$ agreement across all four runs; only $2.09\%$ result in a 3-vs-1 split and $0.84\%$ in a 2-vs-2 tie.
The few criteria with slightly lower 4-run unanimity ($88.3\%$ and $91.5\%$) belong to an open-ended synthetic geometry proof (Question~3) where the model must judge the sufficiency of a student's written geometric argumentation, whereas criteria assessing algebraic equations, coordinates, or exact numerical values in Questions~1, 2, and~4 routinely exceed $96\%$--$100\%$ unanimity.
In short, decomposing grading into explicit binary criteria makes multimodal LLM evaluation highly repeatable, with residual stochasticity restricted to genuinely ambiguous handwritten arguments.

\section{Limitations and Future Work}
\label{sec:limitations}

First, our spatial decomposition strategy relies on cropping each student's answer from a predefined bounding box on the structured examination template.
While this isolates question-level context and prevents cross-question halo effects, an operational limitation arises when a student crosses out their workspace and writes the continuation of their solution outside the designated box, in the margins, or on an unlinked scratch page.
In production deployments, handling such out-of-box responses requires either automated anomaly detection (e.g., flagging scripts where the model detects a pointer to another page or an unexpected zero score) or human review of flagged copies.

Second, although the pipeline automates the evaluation of thousands of scripts, its accuracy remains bounded by the clarity of the additive rubric $\mathcal{C}_q$.
Converting implicit or mixed positive/negative human grading grids into self-contained, strictly positive criteria requires a brief upfront calibration phase by the instructor on a handful of sample copies (4 to 5 copies per question in our study) to verify that alternative valid resolution paths are properly covered.

Third, our evaluation of the generated feedback (RQ3) measures its copy-level specificity rather than its direct pedagogical impact on student learning.
Establishing that the feedback actually helps learners correct their misconceptions requires dedicated follow-up studies—such as measuring instructors' ratings of mathematical accuracy, conducting functional repair tests where students (or simulated student models) revise their flawed solutions using only the feedback, and evaluating perceived usefulness with student cohorts across all questions and subjects.

Finally, the current set of examinations, the students' handwritten answers, and the grading prompts are all in French and focus on university-level mathematics.
While the proposed multimodal grading methodology is language-agnostic, further empirical validation across other languages, educational levels, and non-STEM disciplines is left for future work.

\section{Conclusion}
\label{sec:conclusion}

While the proliferation of generative AI has reinforced the importance of supervised pen-and-paper examinations for assessing authentic student reasoning, manual grading at scale remains constrained by time costs, rater fatigue, sequential contrast effects, and both inter-rater and intra-rater noise. In this paper, we presented and evaluated a stateless multimodal LLM grading methodology that addresses these limitations by evaluating spatially cropped question responses against strictly additive rubric criteria ($w_i > 0$). Through a single schema-constrained model call per question followed by deterministic external score summation, the pipeline decouples semantic evaluation from numerical arithmetic and simultaneously generates auditable examiner justifications and individualized formative feedback for students.

Our empirical evaluation on the June 2026 civil engineering entrance examination in mathematics at UCLouvain ($1{,}752$ handwritten copies across Geometry, Trigonometry, Analysis, and Algebra) demonstrates that frontier multimodal models such as \texttt{gemini-3.8-flash} align closely with multi-grader human reference scores ($r \in [0.92, 0.97]$, $\text{CCC} \in [0.91, 0.96]$, and $\text{MAE}$ between $4.7\%$ and $7.1\%$), actively utilize the full spectrum of rubric criteria, and outperform compact open-weights alternatives (\texttt{glm-5.3-flash}) on complex multi-step derivations. The generated feedback is also specific to each student's answer: an LLM judge recognizes the feedback written for a copy among five candidates in $89.6\%$ of cases (chance level: $20\%$), with residual confusions confined to near-identical, nearly complete solutions. Furthermore, four independent grading runs on the Geometry exam confirm high test-retest stability ($\text{ICC}(2,1) = 0.9912$, mean pairwise difference of $0.58/40$ pts, and $97.1\%$ four-run criterion unanimity). Taken together, these findings indicate that criterion-based multimodal LLM grading can provide reliable, low-cost, and transparent evaluation of complex handwritten STEM assessments while delivering individualized formative feedback at scale.

\clearpage
\appendix

\section{Structured JSON Input and Output for the Illustrative Example}
\label{app:json}

This appendix reports the exact JSON structures exchanged with the multimodal LLM for the illustrative Trigonometry question (\(\sin^2(4x) + \cos^4(2x) = 1\), worth \(8.0\) points) presented in Section~\ref{sec:example}.

\subsection{JSON Schema Enforced at Inference}
To ensure deterministic parsing, the multimodal LLM is constrained via structured output decoding to adhere to the following JSON schema dynamically generated from the question's criterion labels:

\begin{center}
\begin{tikzpicture}
  \definecolor{jsonBoxBdr}{HTML}{CBD5E1}
  \definecolor{jsonBoxBg}{HTML}{F8FAFC}
  \node[rectangle, rounded corners=3pt, draw=jsonBoxBdr, line width=0.65pt, fill=jsonBoxBg,
        inner xsep=10pt, inner ysep=8pt, text width=\dimexpr\linewidth-21.3pt\relax,
        font=\footnotesize\ttfamily, align=left] {%
\{\\
\hspace*{1em}"type": "OBJECT",\\
\hspace*{1em}"properties": \{\\
\hspace*{2em}"JUSTIFICATION": \{ "type": "STRING" \},\\
\hspace*{2em}"FEEDBACK": \{ "type": "STRING" \},\\
\hspace*{2em}"C1\_Double\_Angle": \{ "type": "BOOLEAN" \},\\
\hspace*{2em}"C2\_Factorization": \{ "type": "BOOLEAN" \},\\
\hspace*{2em}"C3\_Solve\_Case1": \{ "type": "BOOLEAN" \},\\
\hspace*{2em}"C4\_Solve\_Case2": \{ "type": "BOOLEAN" \},\\
\hspace*{2em}"C5\_Synthesis\_and\_kZ": \{ "type": "BOOLEAN" \}\\
\hspace*{1em}\},\\
\hspace*{1em}"required": [\\
\hspace*{2em}"JUSTIFICATION", "FEEDBACK",\\
\hspace*{2em}"C1\_Double\_Angle", "C2\_Factorization", "C3\_Solve\_Case1",\\
\hspace*{2em}"C4\_Solve\_Case2", "C5\_Synthesis\_and\_kZ"\\
\hspace*{1em}]\\
\}%
  };
\end{tikzpicture}
\end{center}

\subsection{Raw JSON Response Returned by \texttt{gemini-3.8-flash}}
For the cropped handwritten student answer shown in Figure~\ref{fig:example_crop}, \texttt{gemini-3.8-flash} returns the following raw JSON payload (rendered visually in Figure~\ref{fig:example_output}):

\begin{center}
\begin{tikzpicture}
  \definecolor{jsonBoxBdr}{HTML}{CBD5E1}
  \definecolor{jsonBoxBg}{HTML}{F8FAFC}
  \node[rectangle, rounded corners=3pt, draw=jsonBoxBdr, line width=0.65pt, fill=jsonBoxBg,
        inner xsep=10pt, inner ysep=8pt, text width=\dimexpr\linewidth-21.3pt\relax,
        font=\footnotesize\ttfamily, align=left] {%
\{\\
\hspace*{1em}"JUSTIFICATION": "The student correctly applies the double-angle formula \$\textbackslash sin(4x) = 2\textbackslash sin(2x)\textbackslash cos(2x)\$ to get \$\textbackslash sin\^{}2(4x) = 4\textbackslash sin\^{}2(2x)\textbackslash cos\^{}2(2x)\$ and substitutes it into the original equation (C1 is met). The student rewrites \$1 - \textbackslash cos\^{}4(2x)\$ as \$\textbackslash sin\^{}2(2x)(1 + \textbackslash cos\^{}2(2x))\$ and successfully factors the equation to obtain \$\textbackslash sin\^{}2(2x)[3\textbackslash cos\^{}2(2x) - 1] = 0\$ (C2 is met). Case 1 is solved correctly, yielding \$x = k\textbackslash frac\{\textbackslash pi\}\{2\}\$ with \$k \textbackslash in \textbackslash mathbb\{Z\}\$ (C3 is met). However, in Case 2, the student writes \$\textbackslash cos\^{}2(2x) = \textbackslash frac\{1\}\{3\} \textbackslash iff \textbackslash cos(2x) = \textbackslash frac\{\textbackslash sqrt\{3\}\}\{3\}\$, missing the negative root \$\textbackslash cos(2x) = -\textbackslash frac\{\textbackslash sqrt\{3\}\}\{3\}\$. Consequently, the solutions generated with period \$+k\textbackslash pi\$ miss the negative-root branches (which would have yielded \$\textbackslash pm \textbackslash frac\{1\}\{2\} \textbackslash arccos(\textbackslash frac\{\textbackslash sqrt\{3\}\}\{3\}) + k\textbackslash frac\{\textbackslash pi\}\{2\}\$). Thus, C4 is not met. Finally, the synthesis of the set \$S\$ and the explicit condition \$k \textbackslash in \textbackslash mathbb\{Z\}\$ are properly presented (C5 is met).",\\[3pt]
\hspace*{1em}"FEEDBACK": "The student demonstrated a clear and structured approach to transforming and factoring the trigonometric equation, and solved Case 1 perfectly. However, in Case 2, when taking the square root of \$\textbackslash cos\^{}2(2x) = \textbackslash frac\{1\}\{3\}\$, the student omitted the negative root \$\textbackslash cos(2x) = -\textbackslash frac\{\textbackslash sqrt\{3\}\}\{3\}\$. Taking both signs into account would have yielded the full set of solutions \$x = \textbackslash pm \textbackslash frac\{1\}\{2\} \textbackslash arccos(\textbackslash frac\{\textbackslash sqrt\{3\}\}\{3\}) + k\textbackslash frac\{\textbackslash pi\}\{2\}\$ for \$k \textbackslash in \textbackslash mathbb\{Z\}\$. The student should be cautious not to omit the \$\textbackslash pm\$ when solving equations of the form \$X\^{}2 = c\$.",\\[3pt]
\hspace*{1em}"C1\_Double\_Angle": true,\\
\hspace*{1em}"C2\_Factorization": true,\\
\hspace*{1em}"C3\_Solve\_Case1": true,\\
\hspace*{1em}"C4\_Solve\_Case2": false,\\
\hspace*{1em}"C5\_Synthesis\_and\_kZ": true\\
\}%
  };
\end{tikzpicture}
\end{center}

\clearpage
\section{Additional Empirical Figures}
\label{app:criterion_usage}

\begin{figure}[!ht]
    \centering
    \includegraphics[width=0.95\linewidth]{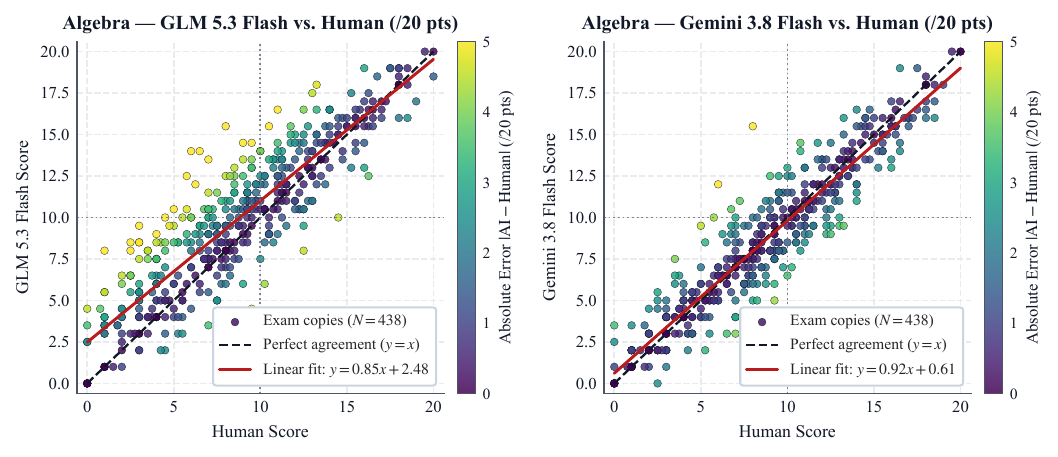}
    \caption{Total score scatter plot and linear regression on the \textbf{Algebra} examination ($N=438$, scale $/20$ pts) for \texttt{glm-5.3-flash} (left) and \texttt{gemini-3.8-flash} (right) against human Gradescope scores.}
    \label{fig:scatter_alg}
\end{figure}

\begin{figure}[!ht]
    \centering
    \includegraphics[width=0.95\linewidth]{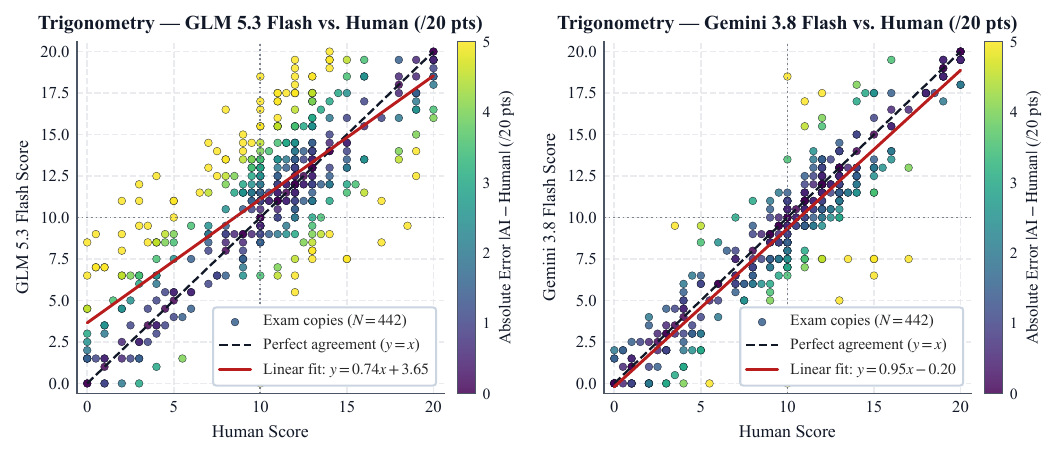}
    \caption{Total score scatter plot and linear regression on the \textbf{Trigonometry} examination ($N=442$, scale $/20$ pts) for \texttt{glm-5.3-flash} (left) and \texttt{gemini-3.8-flash} (right) against human Gradescope scores.}
    \label{fig:scatter_trigo}
\end{figure}

\begin{figure}[!ht]
    \centering
    \includegraphics[width=0.92\linewidth]{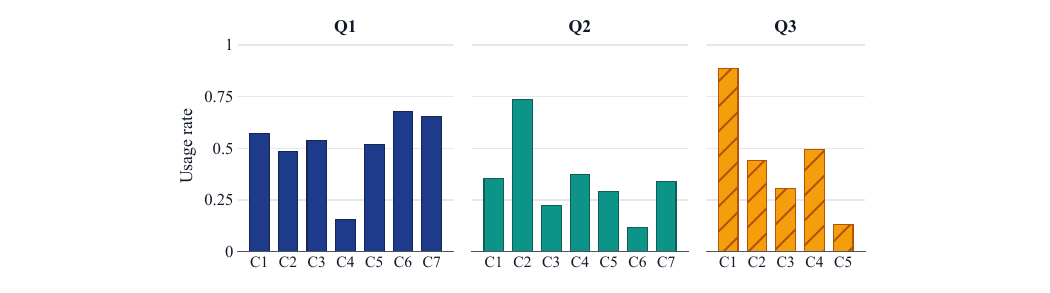}
    \caption{Usage rate of each grading criterion, per question, for the \textbf{Algebra} examination ($N=438$ copies, \texttt{gemini-3.8-flash}).}
    \label{fig:crit_alg}
\end{figure}

\begin{figure}[!ht]
    \centering
    \includegraphics[width=0.92\linewidth]{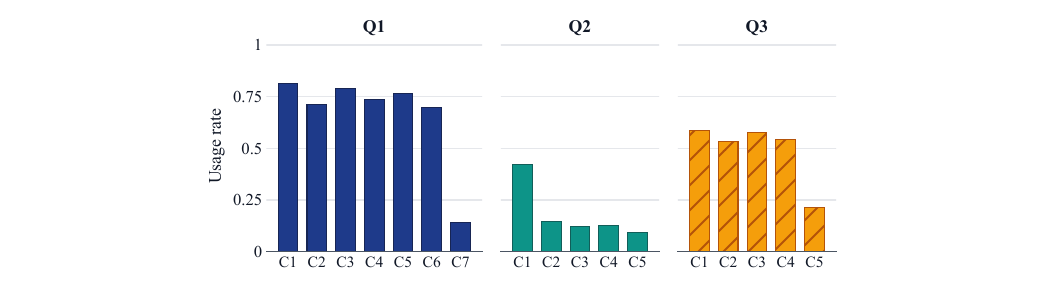}
    \caption{Usage rate of each grading criterion, per question, for the \textbf{Trigonometry} examination ($N=442$ copies, \texttt{gemini-3.8-flash}).}
    \label{fig:crit_trigo}
\end{figure}

\begin{figure}[!ht]
    \centering
    \includegraphics[width=0.92\linewidth]{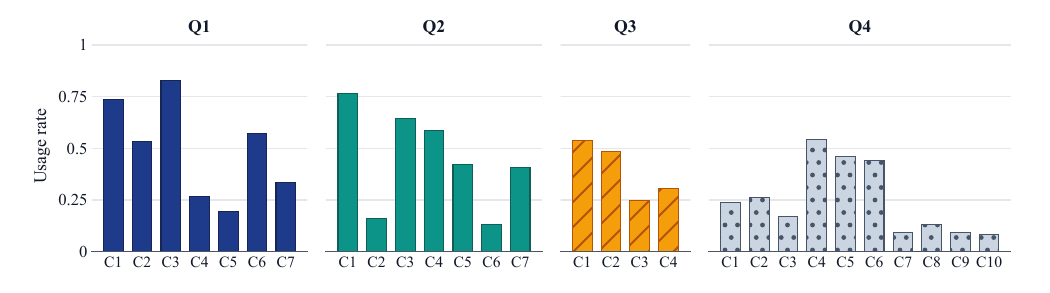}
    \caption{Usage rate of each grading criterion, per question, for the \textbf{Geometry} examination ($N=435$ copies, \texttt{gemini-3.8-flash}).}
    \label{fig:crit_geom}
\end{figure}

\begin{figure}[!ht]
    \centering
    \includegraphics[width=0.98\linewidth]{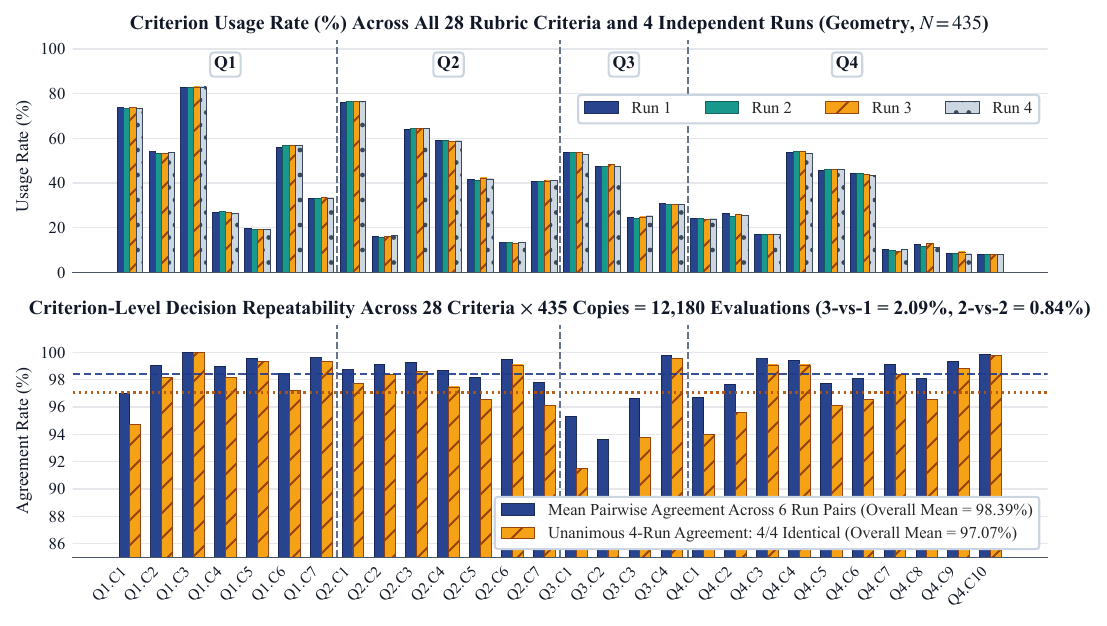}
    \caption{Criterion-level stability across the 4 independent runs on Geometry ($28$ criteria $\times 435$ copies $= 12{,}180$ criterion evaluations per run): usage rate of each criterion across runs (top) and criterion-level decision repeatability showing mean pairwise agreement and 4/4 unanimous agreement (bottom).}
    \label{fig:var_criteria}
\end{figure}

This appendix provides the supplementary figures referenced in Section~\ref{sec:results}: the copy-level total score scatter plots for the Algebra and Trigonometry examinations (Figures~\ref{fig:scatter_alg} and \ref{fig:scatter_trigo}), the criterion usage distributions for Algebra, Trigonometry, and Geometry (Figures~\ref{fig:crit_alg}, \ref{fig:crit_trigo}, and \ref{fig:crit_geom}), and the criterion-level stability across the 4 independent runs on Geometry (Figure~\ref{fig:var_criteria}).

\clearpage
\bibliographystyle{elsarticle-harv}
\bibliography{references}

\end{document}